\pdfoutput=1
\documentclass[letterpaper]{article} % DO NOT CHANGE THIS
\usepackage{aaai24}  % DO NOT CHANGE THIS
\usepackage{times}  % DO NOT CHANGE THIS
\usepackage{helvet}  % DO NOT CHANGE THIS
\usepackage{courier}  % DO NOT CHANGE THIS
\usepackage[hyphens]{url}  % DO NOT CHANGE THIS
\usepackage{graphicx} % DO NOT CHANGE THIS
\usepackage{natbib}  % DO NOT CHANGE THIS AND DO NOT ADD ANY OPTIONS TO IT
\usepackage{caption} % DO NOT CHANGE THIS AND DO NOT ADD ANY OPTIONS TO IT
\usepackage{algorithm}
\usepackage{algorithmic}

\usepackage{verbatim}
\usepackage{amssymb}
\usepackage{multirow}
\usepackage{subfigure}
\usepackage{amsmath}
\usepackage{mathrsfs}

\usepackage{newfloat}
\usepackage{listings}
\DeclareCaptionStyle{ruled}{labelfont=normalfont,labelsep=colon,strut=off} % DO NOT CHANGE THIS
\floatstyle{ruled}
\newfloat{listing}{tb}{lst}{}
\floatname{listing}{Listing}
\title{Question Calibration and Multi-Hop Modeling for Temporal Question Answering}

\author{
    Chao Xue\equalcontrib\textsuperscript{\rm 1},
    Di Liang\equalcontrib\textsuperscript{\rm 2},
    Pengfei Wang\textsuperscript{\rm 3},
    Jing Zhang\textsuperscript{\rm 1}\thanks{Corresponding author.}
}
\affiliations{
    \textsuperscript{\rm 1}School of Software, Beihang University, Beijing, China\\
    \textsuperscript{\rm 2}School of Computer Science, Fudan University, Shanghai, China\\
    \textsuperscript{\rm 3}School of Software, Zhejiang University, Hangzhou, China\\
    \{xuechao, zhang\_jing\}@buaa.edu.cn, liangd17@fudan.edu.cn, wangpf@zju.edu.cn
    
}

\usepackage{bibentry}
\begin{document}

\maketitle

\begin{abstract}
Many models that leverage knowledge graphs (KGs) have recently demonstrated remarkable success in question answering (QA) tasks. In the real world, many facts contained in KGs are time-constrained thus temporal KGQA has received increasing attention. Despite the fruitful efforts of previous models in temporal KGQA, they still have several limitations. (I) They adopt pre-trained language models (PLMs) to obtain question representations, while PLMs tend to focus on entity information and ignore entity transfer caused by temporal constraints, and finally fail to learn specific temporal representations of entities. (II) They neither emphasize the graph structure between entities nor explicitly model the multi-hop relationship in the graph, which will make it difficult to solve complex multi-hop question answering. To alleviate this problem, we propose a novel \textbf{Q}uestion \textbf{C}alibration and \textbf{M}ulti-\textbf{H}op \textbf{M}odeling (\textbf{QC-MHM}) approach. Specifically, We first calibrate the question representation by fusing the question and the time-constrained concepts in KG. Then, we construct the GNN layer to complete multi-hop message passing. Finally, the question representation is combined with the embedding output by the GNN to generate the final prediction. Empirical results verify that the proposed model achieves better performance than the state-of-the-art models in the benchmark dataset. Notably, the Hits@1 and Hits@10 results of QC-MHM on the CronQuestions dataset's complex questions are absolutely improved by 5.1\% and 1.2\% compared to the best-performing baseline. 
Moreover, QC-MHM can generate interpretable and trustworthy predictions.
\end{abstract}

\section{Introduction}
%%%%%%%%%%%%%%%%  cases in chat and google
\begin{figure}[htb]
\centering
\includegraphics[width=0.465\textwidth]{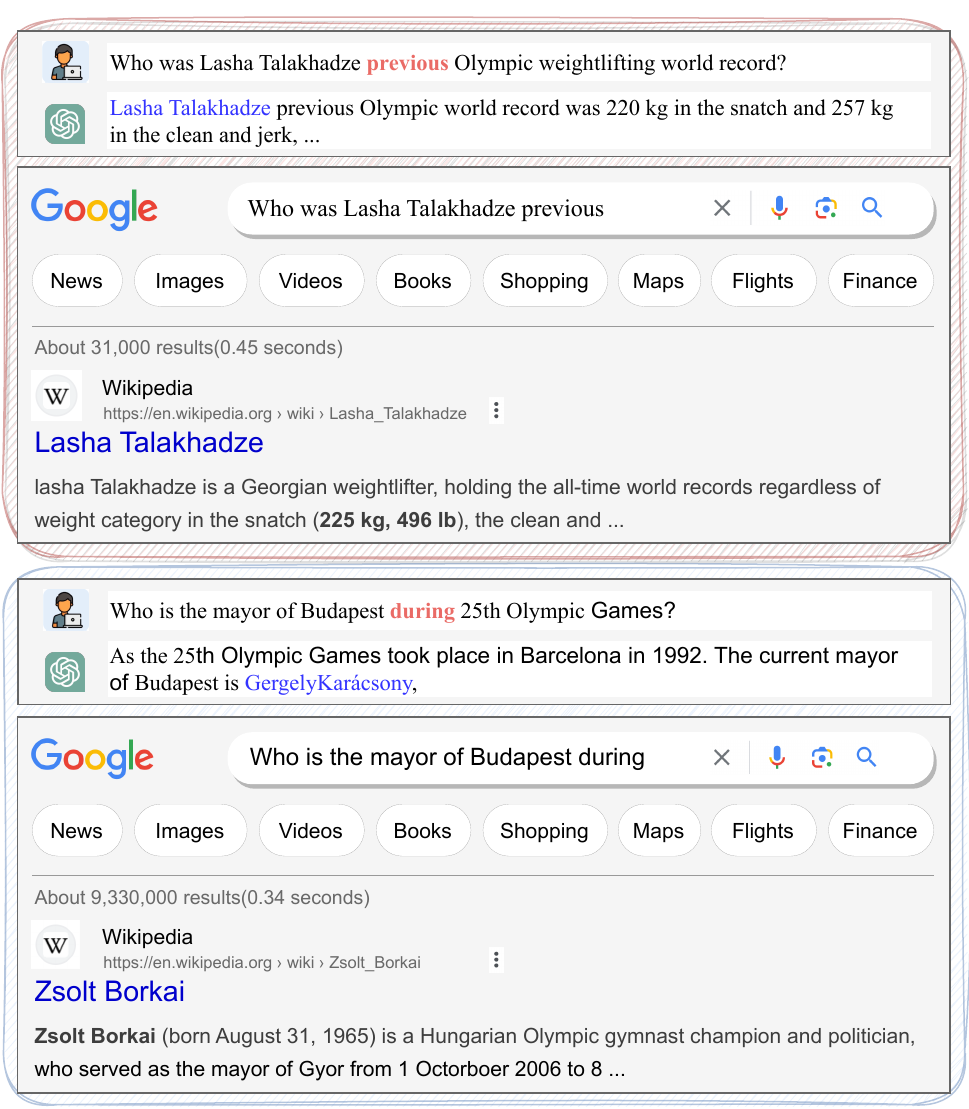}
\caption{\protect\label{fig: example} Examples of complex queries in both Google Search and ChatGPT yield incorrect results.}
\end{figure}

Knowledge graph question answering (KGQA) is a core technique in many NLP applications, such as search and recommendation \cite{huang2019knowledge,xian2019reinforcement,guan2021vpalg}. 
Among several branches of KGQA, temporal KGQA is a recently emerging direction and has shown great potential in real-world practices. There are critical differences between traditional KGQA and temporal KGQA tasks, which are summarized as follows:
(I) Temporal KGQA has more complex semantic information, unlike the traditional KGs constructed based on the tuple of (subject, predicate, object)\footnote{Some researchers refer to predicates as relations. The two are equivalent.}, temporal KGs are attached with additional timestamps.
In other words, the tuple of temporal KGs is (subject, predicate, object, time duration).
(II) Temporal KGQA is expected to generate answers with more diverse types. Different from regular KGQA whose answers are always entities, the answer of temporal KGQA can either be an entity or a timestamp. The above differences make the temporal KGQA more challenging.

To solve the above problem, the limited literature either decomposes the given question into non-temporal and temporal sub-question to answer \cite{jia2018tequila,kingma2014adam} or directly combines the pre-trained language model with the temporal KG to generate answers \cite{saxena2021question,shuman2013emerging}. These methods can achieve satisfying performance on the questions with simple-entity or simple-time (refer to figure~\ref{fig:example} for examples), but fail to answer the questions with complex templates. We argue that the current state-of-the-art methods have not been well solved, and may not even be aware of the following challenges which we address in this paper:

\noindent\textbf{Q1: How to capture implicit or explicit temporal information to calibrate question representation?}
In previous methods, the question is usually encoded by PLMs.
% In other words, they only consider question information and ignore the rich semantic information in the TKG associated with the involved entities. 
The result is that these methods will over-rely on the information of the entities involved in question, and ignore the entity shift caused by time constraints.
Take the question \textit{“Who was lasha talakhadze previous Olympic...”} depicted in Fig.~\ref{fig:example} as an example.
The Google search ignores the time constraint ``previous'' and purely regards ``lasha talakhadze'' and  ``Olympic''  as the query, which leads to the wrong answers. However, there is rich temporal information in temporal KGs, which can promote understanding of the given question. 
%For instance, for the question in Fig.~\ref{fig:example}, one can extract the quadruple (Dani Alves, position held, captain of Brazil, 2019, 2022) in temporal KGs using the entity Dani Alves as the query.
Unfortunately, many prior kinds of research use KGs solely for querying the answer rather than enriching the question representations.

% In essence, these models are retrieval-based approaches that perform well in simple question reasoning. However,

\noindent \textbf{Q2: How to effectively model multi-hop relationships between entities in temporal knowledge graph?}
Existing methods almost don't emphasize the graph structure among entities in temporal KGs and fail to model multi-hop relational paths explicitly, which are not beneficial for reasoning, as demonstrated in previous research [Ren et al., 2020]. Such models struggle when answering the given questions requires multiple facts or multi-hop reasoning (i.e., the second example in Figure \ref{fig:example}). Hence, integrating multi-hop information of KGs can facilitate complex question reasoning, which remains unexplored in temporal KGQA tasks. Moreover, these proposed methods lack certain transparency about their predictions, since they don't model the reasoning paths well and the whole process is invisible.

To address the aforementioned limitations, we propose an \textbf{Q}uestion \textbf{C}alibration and \textbf{M}ulti-\textbf{H}op \textbf{M}odeling (\textbf{QC-MHM}) approach for temporal KGQA. Our goal is to develop a reasoning model that can effectively infer answer entities for the given questions. 
Concretely, for a given question, we first \emph{select} the relevant knowledge (\textit{i.e.}, Subject-Predicate-Object (\textit{abbr}. SPO)) of the entity in the question from TKGs. We design an \emph{question calibration} mechanism to incorporate the SPO into the question representations, which allows the question embeddings to encode the relevant knowledge from temporal KGs, corresponding to \textbf{(Q1)}.
Next, to build the timestamp embeddings with prior knowledge of the temporal order, we employ an auxiliary task for each pair of timestamp embeddings, which is crucial for further improvements in the model performance. Then, to address \textbf{(Q2)}, we explicitly leverage the structural information among entities of KGs via the graph neural networks (GNNs). Moreover, to directly model relational paths, we perform multi-hop message aggregation that allows each node to access its æ-hop neighbors within a single propagation layer, which is significantly superior to one-hop propagation. In modeling relational paths, we introduce an attention mechanism to score the reasoning path. In this way, our model can be interpreted according to this score while reasoning.

% The main contributions of this work can be summarized as follows.
% First, We systematically discuss the feasibility of question representation calibration and multi-hop modeling in temporal KGQA task. And propose a new framework called Question Calibration and Multi-hop Modeling approach. which can better understand the question and infer the correct answer. 
% Secondly, the proposed QC-MHM approach can effectively understand the inherently complex relationship between questions and knowledge graphs, and use the multi-hop module to effectively model multi-hop questions for effective reasoning.
% Finally, Extensive experiments over two temporal datasets demonstrate the superiority of our model compared with other competitive methods. It is worth noticing that, experiments on the CronQuestions dataset, the largest temporal KGQA dataset, the Hits@1 and Hits@10 results of QC-MHM on complex questions are improved by 5.1\% and 1.2\% compared to the best-performing baselines.

\section{Related Work}
\noindent \textbf{Temporal KGQA}:
Generally, KG embedding algorithms \cite{bordes2013translating,yang2014embedding,trouillon2017knowledge} are employed to initialize entity and relation embeddings to help answer a question in the task of KGQA \cite{saxena2020improving}. For temporal KGQA, we typically adopt temporal KG embedding approaches, such as TComplEx \cite{lacroix2020tensor}, for initializing and also obtaining the timestamp embeddings. Recently, many researchers have focused on temporal KGQA and have proposed corresponding methods for this task \cite{jia2018tequila,saxena2021question,mavromatis2021tempoqr,jia2021complex,shang-etal-2022-improving,balcilar2020analyzing}. Among these models, there are three representative ones: CronKGQA \cite{saxena2021question}, TMA \cite{liu2023time,shang2022improving,sharma2022twirgcn} and TSQA \cite{shang2022improving}. CronKGQA utilizes recent advances in temporal KG embeddings and feeds the given questions to pre-trained LMs for answer prediction. Moreover, a dataset, CronQuestions, is proposed in this work, which is larger and more comprehensive than previous benchmarks that mainly employ hand-crafted templates to handle temporal information. Hence, we use CronQuestions as the evaluation dataset. TMA further extracts entity-related information from KGs and adopts a multi-way mechanism for information fusion. TSQA is equipped with a time estimation module that allows unwritten timestamps to be inferred from questions and presents a contrastive learning module that improves sensitivity to time relation words. %However, the above models cannot address sufficiently the limitations previously mentioned, while our model can alleviate them. %The proposed methods are not flexible due to the usage of logical reasoning designed by humans. Although there are previous datasets for this task, they are either too small or require hand-designed templates, so in this paper, we mainly use this dataset because it is more comprehensive compared to the previous ones

\noindent \textbf{Graph Neural Networks}:
GNNs have attracted much attention due to their ability to model structured data and have been developed for various applications in practice \cite{kipf2016semi,velivckovic2017graph,liang2019asynchronous,song-etal-2022-improving-semantic,wang-etal-2022-dabert, zhu2022neural, 
yasunaga2022deep,liang2019adaptive,zheng-etal-2022-robust,hcr:83,garcia2018learning}. Among these models, graph convolutional network\cite{kipf2016semi} is a pioneering work that designs a local spectral graph convolutional layer for learning node embeddings. GraphSAGE \cite{hamilton2017inductive} generates node embeddings by learning an aggregator function that samples and aggregates features from the nodes' local neighborhoods. Graph Attention Network \cite{velivckovic2017graph} assigns different weights to different neighbors of a node to learn its representations by introducing self-attention mechanisms. Recently, several models \cite{feng2020scalable,zhang2018variational,lukovnikov2017neural} have been designed to shift the power of GNNs to general QA tasks. %and achieve satisfying performance. 
However, these models merely use vanilla GNNs that adopt a one-hop neighbor aggregation mechanism, which may limit their expressiveness. %cannot fully utilize their expressive capbilities
Additionally, these models cannot be directly applied to our focused scenarios, \textit{i.e.}, temporal KGQA. %To our best knowledge, we are the first to introduce GNNs to the task of temporal KGQA.

\section{Problem Definition}
\textbf{Temporal KGQA} aims to find suitable answers from KGs $\Omega=(\mathcal{E}, \mathcal{R}, \mathcal{T}, \mathcal{F})$ for given free-text questions. The answer is either an entity or a timestamp. Here, $\mathcal{E}$, $\mathcal{R}$, and $\mathcal{T}$ represent the union sets of entities, relations, and timestamps, respectively. $\mathcal{F}$ denotes the set of facts in the form of quadruples ($s,r,o,t$), where $s, r, o$ and $t$ are the subject, relation, object, and timestamp, respectively.

% Following previous models, we formalize temporal KGQA as a link prediction problem. The underlying idea is to regard the question as a virtual relation to infer the answer. For example, for the question $q$ ``\textit{What award did Cristiano Ronaldo receive in 2008?}'', we can answer it with the single fact (\textit{Cristiano Ronaldo, award received, Ball d'Or}). In fact, we can infer the relation ``award received'' from the question's content, \textit{i.e.}, virtual relation. %get the correct answer by transforming
% Thus, we can solve it by the link prediction manner, which can be transformed into (\textit{Cristiano Ronaldo, $q$, ? 2008}).

%Thus, we can answer it by link prediction, which can be transformed into (\textit{Cristiano Ronaldo, $q$, ?, 2008}).

\noindent \textbf{Temporal KG embeddings} aim to learn low-dimensional embeddings based on the facts contained in the KG. Concretely, we embed $s, o \in \mathcal{E}, r \in \mathcal{R}$ and $t \in \mathcal{T}$ based on the predefined score function $\phi(\cdot)$ to obtain the corresponding embeddings $e_s, e_o, e_r, e_t \in \mathbb{R}^{2D}$. Typically, the valid fact $f=(s,r,o,t)\in \mathcal{F}$ is scored much higher than invalid facts $f^\prime=(s^\prime,r^\prime,o^\prime,t^\prime) \notin \mathcal{F}$, \textit{i.e.}, $\phi(s,r,o,t) \gg \phi(s^\prime,r^\prime,o^\prime,t^\prime)$.

\begin{comment}
TCompIEx \cite{lacroix2020tensor}, a prevailing method specific to temporal KGs, is an extension of CompIEx \cite{trouillon2017knowledge}, which further considers temporal information. Specifically, it is defined in the complex space and its score function is as follows:

\begin{equation}
\begin{aligned}
    \phi(e_s, e_r, \bar{e}_o, e_t)&=\mathbf{Re}(\langle e_s, e_r \odot e_t, \bar{e}_o\rangle) \\
    &=\mathbf{Re}(\sum\nolimits_{d=1}^{2D}e_s[d]e_r[d]e_t[d]\bar{e}_o[d])
\end{aligned}  
\end{equation}
where $\mathbf{Re}$ denotes the real part in the complex space and $\langle \cdot \rangle$ represents the multi-linear product operation. Additionally,  $e_s, e_r e_o, e_t$ are complex-valued embeddings and $\bar{e}_o$ is the complex conjugate of $e_o$. Due to the learning procedure of TcompIEx, it is proficient at inferring missing facts in temporal KGs, such as $(s, r, ?, t)$ and $(s, r, o, ?)$, which is suitable for our scenarios. Therefore, in this work, we combine it with temporal order information to generate pre-trained temporal KG embeddings, which will be described in Section \ref{tim}.
\end{comment}

\section{Proposed Methods}
\label{met}
In this section, we present the details of the QC-MHM. 
which includes three key modules: \textit{time-sensitive KG embedding}, \textit{question calibration and multi-hop modeling}, and \textit{answer prediction}. 
%To better describe the method, we present the overall framework in Fig. \ref{frame}. 
Next, we will elaborate on each module.
%To better understand how it works, its framework is illustrated in Fig. \ref{model}. Concretely, it mainly consists of three essential parts: SPO Selector, Multiway \& Adaptive Fusion, and Answer Prediction. Subsequently, we will elaborate on these critical components that make up our framework.

%In this section, we introduce our proposed model for temporal KGQA, which includes three key modules: \textit{time-sensitive KG embedding}, \textit{information fusion} and \textit{answer prediction}. To better describe the method, we present the overall framework in Fig. \ref{frame}. Next, we will elaborate on each module.

% *********  模型主图   *****************
%\begin{figure*}[ht]%
%    \centering
%    %\scalebox{0.9}{
%    \includegraphics[width=0.9\linewidth]{dafa-new-2 (1).pdf}%}
%    \caption{The overall architecture of our model. (Best viewed in color)}
%    \label{frame}
%\end{figure*}
\subsection{Time-Sensitive KG Embedding}
\label{tim}
We start by obtaining the embeddings of the entity, relation, and timestamp in the temporal KG using a time-sensitive KG algorithm. TComplEx, a prevalent method, can produce high-quality temporal KG embeddings. %is an extension of CompIEx \cite{trouillon2017knowledge}, which further considers the temporal information.
Its score function is as follows:

\begin{equation}
\begin{aligned}
    \phi(e_s, e_r, \bar{e}_o, e_t)&=\mathbf{Re}(\langle e_s, e_r \odot e_t, \bar{e}_o\rangle) \\
    &=\mathbf{Re}(\sum\nolimits_{d=1}^{2D}e_s[d]e_r[d]e_t[d]\bar{e}_o[d])
\end{aligned}  
\end{equation}
where $\mathbf{Re}$ denotes the real part in the complex space and $\langle \cdot \rangle$ represents the multi-linear product operation. Additionally,  $e_s, e_r, e_o, e_t$ are complex-valued embeddings and $\bar{e}_o$ is the complex conjugate of $e_o$. Due to the learning procedure of TComplEx, it is proficient at inferring missing facts in temporal KGs, such as $(s, r, ?, t)$ and $(s, r, o, ?)$, which is suitable for our scenarios. Therefore, in this work, we combine it with temporal order information to generate pre-trained temporal KG embeddings. %, which will be detailed in Section \ref{tim}.

However, the vanilla TComplEx algorithm does not explicitly consider the sequential ordering information of timestamps, which is detrimental to reasoning based on temporal signals. For example, for the question ``\textit{Who was awarded the Ballon d’Or after Lionel Messi?}'', the relevant facts are (\textit{Lionel Messi, award received, Ballon d’Or, [2009, 2009]}) and (\textit{Cristiano Ronaldo, award received, Ballon d’Or, [2013, 2013]}). %(President of USA, position held, Barack. Obama, [2008, 2016]) and (President of USA, position held, Donald. Trump,  [2017, 2021]). 
In the embedding space, it is helpful to be aware that 2013 is later than 2009 when answering this question. Inspired by the usage of position embeddings \cite{vaswani2017attention,jia2021complex,yasunaga-etal-2021-qa,nt2019revisiting,hu2020gpt,liu2022joint,liu2022few,ma2022searching}, we inject temporal order information into timestamp embeddings via an auxiliary task while training temporal KGs. Specifically, we define the position embedding of the $k$-th timestamp $\mathbf{t}_{k}$ as follows:
\begin{equation}
\label{postion}
%\begin{aligned}
    %\mathbf{t}_k(2i)&=\sin(k/10000^{2i/2d}) \\
    %\mathbf{t}_k(2i+1)&=\cos(k/10000^{2i/2d})
    \mathbf{t}_k(c) = \left\{
    \begin{aligned}
         &\sin(k/10000^{2i/2d}), \text{if}\ c=2i \\
         &\cos(k/10000^{2i/2d}), \text{if}\ c=2i+1
    \end{aligned}
    \right.
%\end{aligned}
\end{equation}
where $2d$ is the dimension of timestamps and $c$ denotes the even or odd position in the $2d$-dimensional vector. We can obtain the position embedding $\mathbf{t}_k \in \mathbb{R}^{2d}$ via Eq. \ref{postion}. This position encoding method has the properties of uniqueness (\textit{i.e.}, different timestamps have different position embeddings) and sequential ordering (\textit{i.e.}, it can reflect the relative positions among timestamps).
Next, we adopt linear regression to obtain the probability of timestamp $m$ being ahead of timestamp $n$ for the given pair $(m, n)$. A binary cross-entropy objective function is employed in this auxiliary task. The concrete formulas are as follows:
\begin{equation}
    \begin{aligned}
        \rho(m, n)&=\sigma(\mathbf{W}_{ts}^\top((e_m+\mathbf{t}_m)-(e_n+\mathbf{t}_n))) \\
        \mathcal{L}_{ts}(m,n)&=-\alpha(m,n)\log(\rho(m,n)) \\
        &-(1-\alpha(m,n))\log(1-\rho(m,n))
    \end{aligned}
\end{equation}
where $\sigma(\cdot)$ and $\mathbf{W}_{ts}$ are the sigmoid function and learnable parameters. $e_\ast$ and $\mathbf{t}_\ast$ are timestamp embeddings and the corresponding position embeddings. $\alpha(m, n)$=1 if $m < n$ and 0 otherwise, and $\rho(m,n)$ is the predicted probability of the time order. The final loss function of this module is the weighted sum of the loss function of TComplEx and the auxiliary task, \textit{i.e.}, $\mathcal{L}_{fin}=\mathcal{L}_{tc}+\lambda\mathcal{L}_{ts}$, where $\lambda$ is the weight coefficient and $\mathcal{L}_{tc}$ is the margin loss function as referred to in TComplEx. 
%We can obtain the desired trained embeddings of temporal KGs by performing joint training.

\subsection{Question Calibration and Multi-Hop Modeling}
%This module aims to generate time-aware and multi-hop-aware question representations by combining local information from temporal KGs and semantic information from PLMs.
This module aims to calibrate the question representation by combining semantic information from PLMs and temporal knowledge graphs, while utilizing GNN for multi-hop relation perception.

\noindent \textbf{(I) Question Calibration}:
In order to overcome the entity transfer caused by time constraints, first, we use Sentence-BERT \cite{reimers2019sentence} to find relevant SPO as recall candidates for potential entity transfer to the question. The tokenized question is fed to the sentenceBert to obtain token embeddings. SPO information of temporal KG is also performed in the same operations as above, and we can get the SPO embeddings $S$. The concrete formulas are as Eq. \ref{bert}.
%\begin{footnotesize}
\begin{equation}
\begin{aligned}
    \label{bert}
    Q &= \textrm{sentenceBert}(question) \\
    S &= \textrm{sentenceBert}(<\mathrm{SPO}>)
\end{aligned}
\end{equation}
 %\end{footnotesize}
In general, we take the [CLS] embedding (\emph{i.e.}, $q_{[\mathrm{CLS}]}$ and $q_{s_{[\mathrm{CLS}]}}$) as the final question embedding and SPO embedding.And we apply the cosine similarity on the question and SPO representation to learn the matching scores as follows:
%\begin{footnotesize}
\begin{equation}
    \label{cosine}
    score(q_{[\mathrm{CLS}]}, q_{s_{[\mathrm{CLS}]}}) = \frac{q_{[\mathrm{CLS}]}^\top q_{s_{[\mathrm{CLS}]}}}{\Vert q_{[\mathrm{CLS}]}\Vert \Vert q_{s_{[\mathrm{CLS}]} \Vert}}
\end{equation}
%\end{footnotesize}
\noindent where $score$ is a scalar. The top ten scored SPOs are selected as candidate information. Then, previous studies \cite{rocktaschel2015reasoning,ijcai2018p613, xue2023dual} demonstrate the effectiveness of word-level attention in sentence pair modeling. Inspired by this, in order to model the relationship between questions and SPO from different perspectives, we design a multi-view alignment module, which uses different types of attention functions to compare the correlation of questions and SPO from different perspectives.

For a given question, we embed it with Eq. \ref{bert}, excluding the [CLS] token, \emph{i.e.}, $\bar{Q}=[q_1, q_2, \dots, q_n]$. For the ten selected SPOs, we take the [CLS] token of each SPO and concatenate them together, \emph{i.e.}, $P=[S_1, S_2, \dots, S_{m}]$ ($m$ is the number of selected SPOs). Then, the candidate SPOs can be matched by the word at each position $k$ of the question. %This procedure is based on three attention mechanisms, namely, concat attention, dot attention, and minus attention, %to obtain corresponding weighted-sum representations of $\tilde{p}_k^\ell$. 
which are formulated as follows:
%\begin{footnotesize}
\begin{equation}
    \tilde{p}_k^\ell = \Phi_\ell(P, q_k;\mathbf{W}_\ell)
\end{equation}
%\end{footnotesize}
\noindent where $\tilde{p}_k^\ell$ is the corresponding weighted-sum representation of SPOs specified by $q_k$, employing the attention function $\Phi_\ell$ with parameterized by $\mathbf{W}_\ell$, in which $\ell$ denotes concat attention, dot attention, and minus attention, respectively. More precisely, the different attention mechanisms can be described as follows:

\noindent \textbf{Concat Attention}:
\begin{footnotesize}
\begin{equation}
\begin{aligned}
    h^k_j&=v_{cat}^\top\tanh(\mathbf{W}_{cat}[q_k, S_j]) \\
    \alpha_i^k&=\exp(h^k_i)/\sum\nolimits_{j=1}^m\exp(h^k_j) ,\quad \tilde{p}_k^{cat}=\sum\nolimits_{i=1}^m\alpha_i^kS_i
\end{aligned}
\end{equation}
\end{footnotesize}

\noindent \textbf{Dot Attention}:
\begin{footnotesize}
\begin{equation}
\begin{aligned}
    h^k_j&=v_{dot}^\top\tanh(\mathbf{W}_{dot}(q_k \odot S_j)) \\ \alpha_i^k&=\exp(h^k_i)/\sum\nolimits_{j=1}^m\exp(h^k_j),\quad \tilde{p}_k^{dot}=\sum\nolimits_{i=1}^m\alpha_i^kS_i
\end{aligned}
\end{equation}
\end{footnotesize}

\noindent \textbf{Minus Attention}:
\begin{footnotesize}
\begin{equation}
\begin{aligned}
    h^k_j&=v_{min}^\top\tanh(\mathbf{W}_{min}(q_k - S_j))\\ \alpha_i^k&=\exp(h^k_i)/\sum\nolimits_{j=1}^m\exp(h^k_j),\quad \tilde{p}_k^{min}=\sum\nolimits_{i=1}^m\alpha_i^kS_i
\end{aligned}
\end{equation}
\end{footnotesize}

%\noindent where $v^\top_*$ and $\mathbf{W}_*$ are all learnable parameters. $h_j^k$ and $\alpha_i^k$ denote hidden representations and the normalized attention coefficients, respectively.

Next, to obtain the attention-based question representation $\tilde{Q}^\ell$, we aggregate the matching information $\tilde{p}_k^\ell$ together with the word representation $q_k$ via the concatenation operation, \textit{i.e.}, $\tilde{q}_k^\ell=[q_k, \tilde{p}_k^\ell]$. Finally, the linear transformation is applied to $\tilde{Q}^\ell$ that fuses the SPO information, \emph{i.e.}, $\mathcal{Q}_{final}=\mathbf{W}[\tilde{Q}^{cat}, \tilde{Q}^{dot}, \tilde{Q}^{min}]=[\hat{q}_1, \dots, \hat{q}_n]$. Similarly, the question can be matched by a particular SPO by performing the above multiway operation and linear transformation. In this way, we can obtain the updated SPO representation $\hat{S}_i$.

Finally, we use adaptive fusion to make question representations more time-aware, we use a gate mechanism to adaptively fuse the temporal information from SPOs.
%\begin{footnotesize}
\begin{equation}
    \begin{aligned}
        %\hat{q}_s = \tanh(\mathbf{W}_{\hat{q}_s}q_s+b_{\hat{q}_s}) \\
        \tilde{S} &= \tanh(\mathbf{W}_{\hat{S}_i}\frac{1}{m}\sum\nolimits_{i=1}^m\hat{S}_i+b_{\hat{S}_i}) \\
        g_i &= \sigma(\mathbf{W}_{g_i}(\hat{q}_i\cdot\tilde{S})) ,\quad q_{\mathrm{sem}}=g_i\hat{q}_i + (1-g_i)\tilde{S}
    \end{aligned}
\end{equation}
%\end{footnotesize}
\noindent where $\sigma$ denote the nonlinear activation function, respectively. $q_{\mathrm{sem}}$ is the final embedding of the word in the question, which is the representation containing the potential entity transfer information.

\noindent \textbf{(II) Multi-Hop Modeling}: We first construct a graph $G=(V, E)$ based on given temporal KGs, where $V$ is the set of nodes, denoting entities, and $E$ is the set of edges, connecting the triplet's subject and object. The value of an edge is the concatenation of a relation and timestamp, \textit{i.e.}, $r||t$. The idea is to propagate both relations and timestamps via graph structures, which is specific to temporal KGQA tasks. The node and edge features can be initialized by the pre-trained time-sensitive KG encoder. 

Next, we obtain annotated entities $\{\text{ent}_1, \text{ent}_2,\cdots, \text{ent}_w\}$ from each question $q$. For each entity $\text{ent}_i$, we then extract its \ae-hop sub-graph $G_i$. The final relevant \ae-hop sub-graph $G_q$ for the question can be obtained by combining each entity's sub-graph, \textit{i.e.}, $G_q=\cup_{i=1}^{w}G_i$. Note that we restrict the answer selection to $G_q$ via the latent sub-graph extraction procedure, which can greatly reduce the search space and effectively facilitate the training process.

To directly leverage the structural information among entities of temporal KGs, we apply GNNs to the extracted sub-graph. %GNNs are applied to the extracted subgraph. 
Typically, the classic message passing paradigm of GNNs can be formulated as:
\begin{equation}
\begin{aligned}
    a_v^\ell &= \textbf{AGGREGATE}(\{h_u^{\ell-1}:u \in \mathcal{N}_v\}) \\
    h_v^\ell &= \textbf{COMBINE}(h_v^{\ell-1}, a_v^\ell)
\end{aligned}
\end{equation}
where $\mathcal{N}_v$ is the set of node $v$'s neighbors. $a_v^\ell$ is the aggregated message at layer $\ell$, and $h_v^\ell$ is node $v$'s embeddings at layer $\ell$ obtained by combining $h_v^{\ell-1}$ and $a_v^\ell$. However, in the above framework, the nodes in the graph can only access their one-hop neighbors through a single graph layer. In other words, suppose two nodes are not directly connected, they can only interact with each other by stacking a sufficient number of layers, which severely limits the capability of GNNs to explore the relationships between disjoint nodes. %More details as shown in Appendix \ref{display}. %Fig. \ref{layer}. 

To address this problem, we adopt a multi-hop message passing mechanism that works on all possible paths between two nodes. The first step is to compute the normalized attention using Eq. \ref{attention}.%allows for influence between disjoint nodes.
%For the question, ``What's the relationship between Mical and Brack when ''the one-hop message passing limits their ability to explore the relationship between the broadergraph structure.
\begin{equation}
\label{attention}
    \begin{aligned}
        \mathscr{A}_{irj}^\ell&=\left\{
        \begin{aligned}
        \beta^\ell\delta&(\mathbf{W}_{ad}^\ell(h_i^\ell||h_j^\ell||h_r||h_t)), (v_i, r, v_j, t) \in G \\
        &-\infty, \text{otherwise}
        \end{aligned}
        \right. \\
        \mathbf{A}^\ell &= \text{softmax}(\mathscr{A}^\ell) 
    \end{aligned}
\end{equation}
where $\beta^\ell$ and $\mathbf{W}_{ad}^\ell$ are the learnable weights shared by the $\ell$-th layer. $h_i^{\ell}$ is the embedding of node $i$, initialized by $h_{i}^0=e_i$. $h_r$ and $h_t$ are the embeddings of relation $r$ and timestamp $t$, respectively. $\delta$ denotes the ReLU activation function. $\mathscr{A}^\ell$ and $\mathbf{A}^\ell$ represent the attention matrix obtained by applying the edges appearing in $G$ and the normalized attention matrix derived by performing a row-wise softmax function, respectively. In addition, since paths with different importance are assigned corresponding weights using Eq. \ref{attention}, we can derive the reasoning path based on these weights.
$||$ denotes the concatenation operation. To enable the aggregation of multi-hop messages to a target node within a single propagation layer, we employ a mechanism defined as follows:
\begin{equation}
\label{matrix}
    \mathbf{D}=\sum\nolimits_{\tau=0}^{\aleph}\xi_\tau\mathbf{A}^{(\tau)}
\end{equation}
where $\xi_\tau$ are trainable vectors. $\mathbf{A}^{(\tau)}$ is the powers of $\mathbf{A}$, which considers relational paths with length limits up to $\tau$ from neighboring nodes to the target node. In other words, the target node's context (\textit{i.e.}, intermediate neighbors) and its local graph structure are involved in attention calculation. This procedure successfully creates attentional interactions between a node and its disjoint neighbors beyond one-hop. In practice, we can achieve impressive performance when empirically setting the diffusion distance $\aleph \in [2, 4]$ since many graphs have small-world properties with lower diameters. Subsequently, the transition matrix $\textbf{D}$ is leveraged to update the nodes' embeddings to obtain $\mathbf{H}^{\ell+1}$ in Eq. \ref{update}.
\begin{equation}
\label{update}
    \mathbf{H}^{\ell+1}=\mathbf{D}\mathbf{\mathbf{H^\ell}}
\end{equation}
%To make the obtained features more informative, we use the multi-head mechanism. 
Finally, we perform an average pooling operation on the nodes of the extracted sub-graph to acquire the question's multi-hop information $\mathbf{Q}_{mlh}$, formulated as Eq. \ref{ba}.
\begin{equation}
\label{ba}
    \mathbf{Q}_{mlh}=\frac{1}{|V_q|}\sum\nolimits_{i\in V_q}h^L_i
\end{equation}
where $V_q$ is the node set of the sub-graph and $h^L_i$ is the node embeddings at the $L$-th layer.

%Although previous GNN models feature multi-hop message passing mechanisms, they either focus on single-relation graphs \cite{liu2021deep} or set stricter constraints \cite{wang2020multi}, which limits their expressiveness. However, the multi-hop propagation we employ can be applied to multi-relational graphs (\textit{e.g.}, KGs) and does not set greater constraints on the coefficients (\textit{i.e.}, $\xi_\tau$). 
%Moreover, we prove \textbf{theoretically}  that multi-hop propagation is more expressive than one-hop propagation by applying spectral graph analysis, and the full proof can be found in \textbf{Appendix A.2}. %\ref{app}.

To better integrate the question's local and global information, we employ a sophisticated knowledge fusion layer $\Phi(\cdot)$, that contains several Transformer encoder layers. After performing the  Transformer-based information fusion layer, we obtain the final question representation, \textit{i.e.}, $\mathbf{Q}_{fin}=\Phi(\mathbf{Q}_{sem}||\mathbf{Q}_{mlh})$.

\subsection{Answer Prediction}
We use two-layer MLPs to transform $\mathbf{Q}_{fin}$ into $\mathbf{Q}_{ent}$ and $\mathbf{Q}_{tim}$, which correspond to entity and timestamp prediction, respectively and are defined in Eq. \ref{transform}.
\begin{equation}
\label{transform}
\begin{aligned}
    \mathbf{Q}_{ent}=\mathbf{MLP}(\mathbf{Q}_{fin}) \\
    \mathbf{Q}_{tim}=\mathbf{MLP}(\mathbf{Q}_{fin})
\end{aligned}
\end{equation}
Next, we define an entity score function $\phi_{ent}(\cdot)$ and a timestamp score function $\phi_{tim}(\cdot)$ to obtain the scores of candidate entities and timestamps, as shown in Eq. \ref{score}.

\begin{equation}
\label{score}
\begin{aligned}
     \phi_{ent}(\tilde{\varepsilon})=\mathbf{Re}(\langle e_s, \mathbf{Q}_{ent}\odot e_t, \bar{e}_{\tilde{\varepsilon}} \rangle) \\
     \phi_{tim}(\tilde{t})=\mathbf{Re}(\langle e_s, \mathbf{Q}_{tim}\odot e_{\tilde{t}},\bar{e}_o 
     \rangle)
\end{aligned}
\end{equation}
where $\tilde{\varepsilon} \in \mathcal{E}_q$ and $\tilde{t} \in \mathcal{T}_q$, in which $\mathcal{E}_q \subseteq \mathcal{E}$ and $\mathcal{T}_q \subseteq \mathcal{T}$ are specified by the sub-graph $G_q$ with respect to the given question $q$. 

Finally, we concatenate the obtained scores for the entities and timestamps and perform the softmax function over them to obtain the answer probability. The objective function is the cross-entropy loss, as shown in Eq. \ref{train}.
\begin{equation}
\label{train}
\begin{aligned}
     \hat{y}_i=\text{softmax}(\phi_{ent}(\cdot)||\phi_{tim}(\cdot)) \\
     \mathcal{L}_{predict}=-\sum\nolimits_iy_i\log(\hat{y}_i)
\end{aligned}
\end{equation}
where $y_i$ is the true answer to the question.

\begin{table*}[h]
	\centering
 \renewcommand\arraystretch{1.08}

	\scalebox{0.95}{
        \setlength{\tabcolsep}{2.5mm}{
	\begin{tabular}{l|c|cc|cc|c|cc|cc}
	    \hline
		~& \multicolumn{5}{c|}{\textbf{Hits@1}} & \multicolumn{5}{c}{\textbf{Hits@10}} \\
		\cline{2-11}
        {\textbf{Model}}& {\textbf{Overall}} & \multicolumn{2}{c|}{\textbf{Question Type}} & \multicolumn{2}{c|}{\textbf{Answer Type}} & {\textbf{Overall}}& \multicolumn{2}{c|}{\textbf{Question Type}} & \multicolumn{2}{c}{\textbf{Answer Type}} \\
        \cline{3-6}\cline{8-11}
        ~&~& \textbf{Complex} & \textbf{Simple} & \textbf{Entity}&\textbf{Time}&~&\textbf{Complex} & \textbf{Simple} & \textbf{Entity}&\textbf{Time}\\
        \hline
        BERT & 0.243 & 0.239 & 0.249 & 0.277 & 0.179 & 0.620 & 0.598 & 0.649 & 0.628 & 0.604 \\
        RoBERTa & 0.225 & 0.217 & 0.237 & 0.251 & 0.177 & 0.585 & 0.542 & 0.644 & 0.583 & 0.591 \\
        KnowBERT & 0.226 & 0.220 & 0.238 & 0.252 & 0.177 & 0.586 & 0.539 & 0.646 & 0.582 &0.592\\
        T5-3B&0.252 & 0.240 &0.251 & 0.283 & 0.180&-&-&-&-&- \\
        \hline
        EmbedKGQA & 0.288 & 0.286 & 0.290 & 0.411 & 0.057 & 0.672 & 0.632 & 0.725 & 0.85& 0.341\\
        T-EaE-add& 0.278 & 0.257 & 0.306 & 0.313& 0.213& 0.663 &0.614 & 0.729 & 0.662 & 0.665\\
        T-EaE-replace & 0.288 & 0.257 & 0.329 & 0.318 & 0.231 & 0.678 & 0.623 & 0.753 & 0.668 & 0.698\\
        \hline
        CronKGQA & 0.647 & 0.392 & 0.987 &0.699 & 0.549 & 0.884&0.802&0.992 & 0.898 & 0.857\\
        TMA  & 0.784 & 0.632 & 0.987 & 0.792 & 0.743 & 0.943 & 0.904 & 0.995 & 0.947 & 0.936\\
        TSQA  & 0.831 & 0.713 & 0.987	& 0.829	& 0.836	& 0.980	& 0.968	& 0.997	& 0.981	& 0.978 \\
        TempoQR  & 0.918 & 0.864 & 0.990 & 0.926 & 0.903 & 0.978 & 0.978 & 0.993 & 0.980 & 0.974\\
        CTRN  & 0.920 & 0.869 & 0.900 & 0.921 & 0.917 & 0.980 & 0.970 & 0.993 & 0.982 & 0.976\\
        QC-MHM (ours)& \textbf{0.971} & \textbf{0.946} & \textbf{0.992}	& \textbf{0.962} & \textbf{0.966} & \textbf{0.992} & \textbf{0.986}	& \textbf{0.998} & \textbf{0.993} & \textbf{0.987} \\
        \hline
        Ab. Imp & 5.1\% & 7.7\% & 0.2\% & 3.6\% & 4.9\% & 1.2\% & 1.6\% & 0.1\% & 1.1\% & 0.9\% \\\hline
	\end{tabular}}}
 \caption{Performance of baselines and our methods on the CronQuestions dataset.}%, please refer to Section~\ref{res:MP} for details.}
	\label{results} 
\end{table*}
% \vspace{-0.5cm}

\begin{table}[h]
	\centering
  \renewcommand\arraystretch{1.08}
	
	% \scalebox{0.96}{
	\resizebox{0.48\textwidth}{!}{
	\begin{tabular}{l|c|c c|c c}
	    \hline

        {\textbf{Model}}& {\textbf{Overall}} & {\textbf{Explicit}} & {\textbf{Implicit}} & {\textbf{Temporal}} & {\textbf{Ordinal}} \\

        \hline
        CronKGQA &0.393 &0.388 &0.380 &0.436 &0.332	\\
        TMA  &0.436 &0.442 &0.419 &0.476 &0.352 \\
        TempoQR &0.459 &0.503 &0.442 &0.458 &0.367\\
        CTRN  &0.465 &0.469 &0.446 &0.512 &0.382\\
        QC-MHM  &0.531 &0.533 &0.508 &0.607 &0.401\\
        \hline
	\end{tabular}
	}
 \caption{Hits@1 for different models on TimeQuestions.}
\label{ablation2}  
	% }
\end{table}
% \vspace{-0.5cm}

\section{{Datasets and Baselines}}
\subsection{Datasets}
We employ two temporal KGQA benchmarks,
CRONQUESTIONS \cite{saxena2021question} and Time-Questions \cite{jia2021complex}.
\textbf{CRONQUESTIONS} is the largest known dataset and is widely used in previous studies. It contains two main parts: the Wikidata temporal KG and a collection of free-text questions. Among them, there are 125K entities, 1.7K timestamps, 203 relations, and 328K facts in the form of quadruples in the temporal KG. Additionally, 410K unique question-answer pairs are given, where each question contains annotated entities and timestamps. Moreover, this dataset can be divided into entity and time questions based on the type of answers. It can also be divided into simple reasoning (\textit{i.e.}, Simple Entity and Simple Time) and complex reasoning (\textit{i.e.}, Before/After, First/Last, and Time Join) based on the questions' difficulty.  
\textbf{TimeQuestions} is another challenging dataset, which has
16k manually tagged temporal questions and are divided into
four categories (\textit{i.e.,} Explicit, Implicit, Temporal, and Ordinal) according to the type of time reasoning. 
%\noindent \textbf{Datasets}:

\subsection{Baselines} 

\noindent \textbf{Pre-Trained LMs} 1) BERT\cite{devlin2018bert}: It employs the bidirectional Transformer to encode a large-scale corpus. In our experiments, we use BERT to produce the question embeddings and concatenate them with pre-trained entity and timestamp embeddings to predict the answer. 2) RoBERTa\cite{liu2019roberta}: It extends BERT by using a dynamic mask. Moreover, it is trained with a larger batch size and corpus. To answer each given question, we utilize RoBERTa to generate question embeddings and concatenate them with pre-trained entity and timestamp embeddings. 3) KnowBERT \cite{peters2019knowledge}: It is a variant of BERT that introduces information from KGs, such as WikiData and WordNet. The [CLS] token is adopted in answer prediction. 4) T5 \cite{raffel2020exploring}: It proposes a powerful architecture called Text-to-text. To apply T5 to temporal questions, we transform each question into the form ``temporal question: question?''.

\noindent \textbf{General KG Embedding-Based Models} 1) EaE \cite{fevry2020entities}: It is an entity-aware method integrating entity knowledge into Transformer-based LMs. 2) EmbedKGQA \cite{saxena2020improving}: It leverages the CompIEx method to produce KG embeddings for KGQA tasks and can only deal with non-temporal questions. To accommodate this task, random timestamp embeddings are used during the training stage.

\noindent \textbf{Temporal KG Embedding-Based Models} 1) CronKGQA \cite{saxena2021question}: It further extends EmbedKGQA by employing temporal KG embeddings and is designed for TKGQA tasks.
2) TMA \cite{liu2022time,fei2022cqg,gui2018transferring}: It extracts the relevant SPO information and adopts multi-way attention to enhance question understanding.
3) TSQA \cite{shang2022improving}: It proposes a time-sensitive module to infer the time and contrastive losses to improve the model's ability to perceive time relation words, achieving promising performance.

\section{Results}
%\textbf{Model Performance}:
\subsection{Model Performance}
We present the results of our proposed QC-MHM and baselines on CRONQUESTIONS in terms of Hits@1 and Hits@10 in Table \ref{results}. QC-MHM achieves the best performance in all experimental settings, indicating its superiority on the temporal KGQA task. Remarkably, QC-MHM significantly outperforms the second-best model on both datasets. It achieves 7.7\% and 4.9\% absolute improvements on Hits@1 with respect to complex reasoning and time questions on CRONQUESTION, respectively. It also performs far better than other models for various types of questions in the TimeQuestions dataset in Table \ref{ablation2}. 
For example, it achieves absolute improvements of 3.0\% and 6.2\% on Hits@1 for questions involving ‘Explicit’ and ‘Implicit’ types.
While in the ‘Temporal’ type of questions, an absolute improvement of 9\% is obtained over the best baseline.
%We attribute this to the use of the multi-hop propagation of knowledge fusion and the time-sensitive KG embedding.

\begin{table}[h]
	\centering
 \renewcommand\arraystretch{1.08}

	\resizebox{0.48\textwidth}{!}{
	\begin{tabular}{l|c c c|c c|c}
	    \hline
	    ~&\multicolumn{3}{c|}{\textbf{Complex Question}}&\multicolumn{2}{c|}{\textbf{Simple Question}}& \\
	    \cline{2-6}
        {\textbf{Category}}&\textbf{Before/}&\textbf{First/}&\textbf{Time}&\textbf{Simple}&\textbf{Simple}& {\textbf{All}} \\
        ~&\textbf{After} & \textbf{Last} & \textbf{Join} & \textbf{Entity}& \textbf{Time} & ~ \\
        \hline
        EmbedKGQA & 0.199 & 0.324 & 0.223 & 0.421 & 0.087 & 0.288 \\
        T-EaE-add& 0.256 & 0.285 & 0.175 & 0.296 & 0.321 & 0.278\\ 
        T-EaE-replace & 0.256 & 0.288 & 0.168 & 0.318 & 0.346 & 0.288\\ 
        CronKGQA&0.288 & 0.371 &0.511 & 0.988 & 0.985&0.647 \\
        TMA &0.581 &0.627&0.675&0.988&0.987&0.784 \\ 
        TSQA &0.504 &0.721 &0.799&0.988&0.987 &0.831\\
        TempoQR &0.714 &0.853 &0.978&0.988&0.987 &0.918\\
        CTRN &0.747 &0.880 &0.897&0.991&0.987 &0.920\\
        QC-MHM &\textbf{0.905} & \textbf{0.938} &\textbf{0.992} &\textbf{0.989} &\textbf{0.996} & \textbf{0.971} \\
        \hline
	\end{tabular}
	}
	\caption{Hits@1 for different question types.}
 	\label{category}
\end{table}
% \vspace{-0.5cm}

\begin{table}[h]
	\centering
 \renewcommand\arraystretch{1.08}
	%\scalebox{0.6}{
	\resizebox{0.48\textwidth}{!}{
	\begin{tabular}{l|c|c c|c c}
	    \hline
	    ~& \multicolumn{5}{c}{\textbf{Hits@1}}\\
		\cline{2-6}
        {\textbf{Model}}& {\textbf{Overall}} & \multicolumn{2}{c|}{\textbf{Question Type}} & \multicolumn{2}{c}{\textbf{Answer Type}} \\
        \cline{3-6}
        ~&~& \textbf{Complex} & \textbf{Simple} & \textbf{Entity}&\textbf{Time}\\
        \hline
        QC-MHM &0.971	&0.946	&0.992	&0.962	&0.966\\
        w/o Time Order &0.902	&0.884	&0.949	&0.883	&0.916\\
        w/o Multi-Hop &0.868	&0.841	&0.903	&0.854	&0.909\\
        w/o Calibration &0.826	&0.753	&0.872	&0.738	&0.871\\
        \hline
	\end{tabular}
	%}
	}
	\caption{Results of ablation studies on our model.}
	\label{ablation}  
\end{table}
% \vspace{-0.5cm}

We find that PLMs (\textit{e.g.}, BERT) achieve unsatisfactory performance in this scenario, lagging far behind the TKG embedding-based models. A plausible reason is that these models do not introduce KG into this task, which is detrimental to question understanding. Despite the relative success of general KG embedding-based models (\textit{e.g.}, EmbedKGQA) in common QA tasks, they still perform worse than TKG embedding-based models(\textit{e.g.},TMA, and TSQA) in our focused scenario. A possible reason is that they do not explicitly leverage temporal KG and neglect temporal information, which is crucial for the temporal KGQA task.

We present the Hits@1 results of our model and other baselines on different types in Table \ref{category}. QC-MHM is significantly superior to other models, especially for complex questions. Our model gains 15.8\%, 5.8\%, and 1.4\% absolute improvement over ``Before/After'', ``First/Last'' and ``Time Join'', respectively, due to the consideration of the timestamp order and multi-hop structural information of the TKG. 
%Additionally, our model has comparable performance on simple questions.

\begin{figure}
    \centering
    \includegraphics[width=0.43\textwidth]{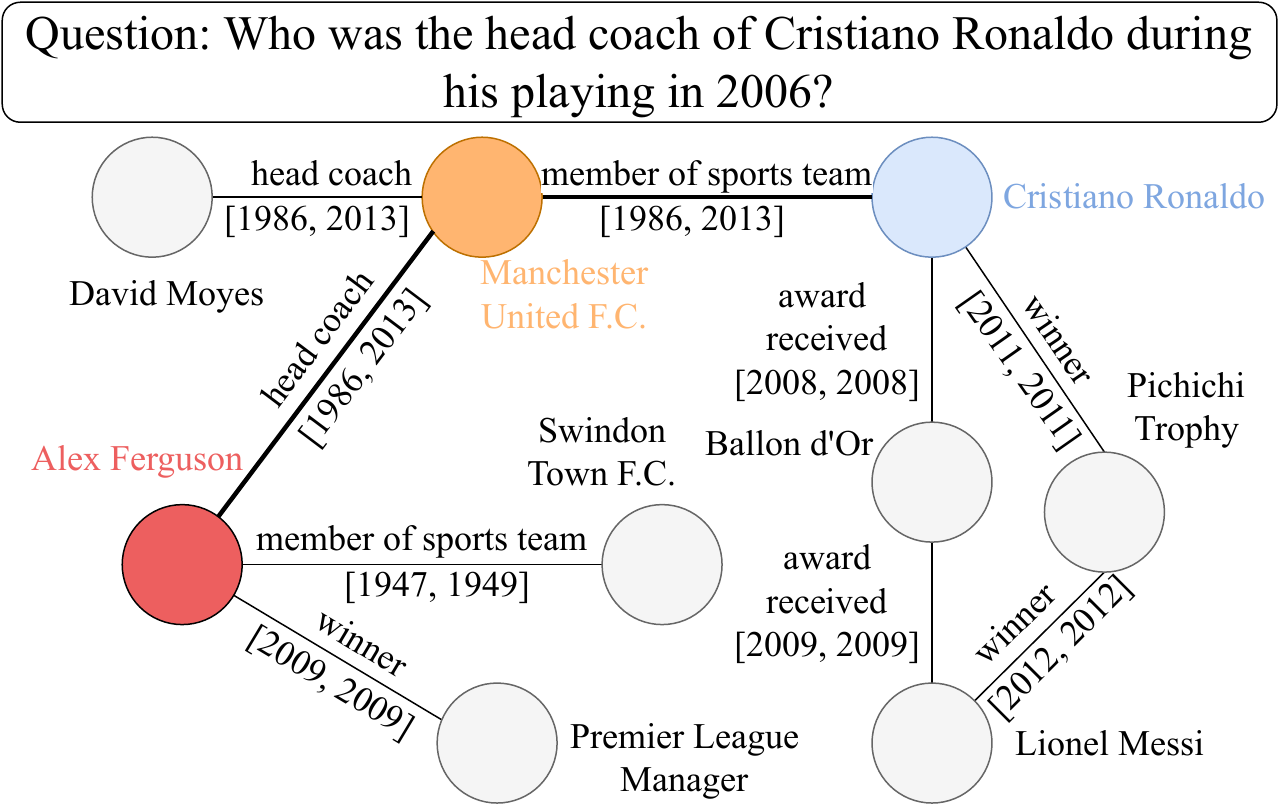}
    \caption{Visualization of a case study of the interpretability of our model. For brevity, we only show the key entities.}
    \label{example}
\end{figure}

%\textbf{Ablation Study}:
\subsection{Ablation Study}
To evaluate the contribution of each module in our framework, we perform extensive ablation experiments. The experimental results are shown in Table \ref{ablation}.
%We conduct extensive ablation experiments on the crucial components by designing some model variants.
(I) \textit{W/o Time Order}: We exclude temporal order encoding and use the vanilla TComplEx method. 
(II) \textit{W/o Multi-Hop}: We use the one-hop attention computed from the neighbors without multi-hop attention, similar to GAT. 
(III) \textit{W/o Question Calibration}: We removed the module of Question Calibration.

\begin{figure}[bp]
    \centering
    \includegraphics[width=0.48\textwidth]{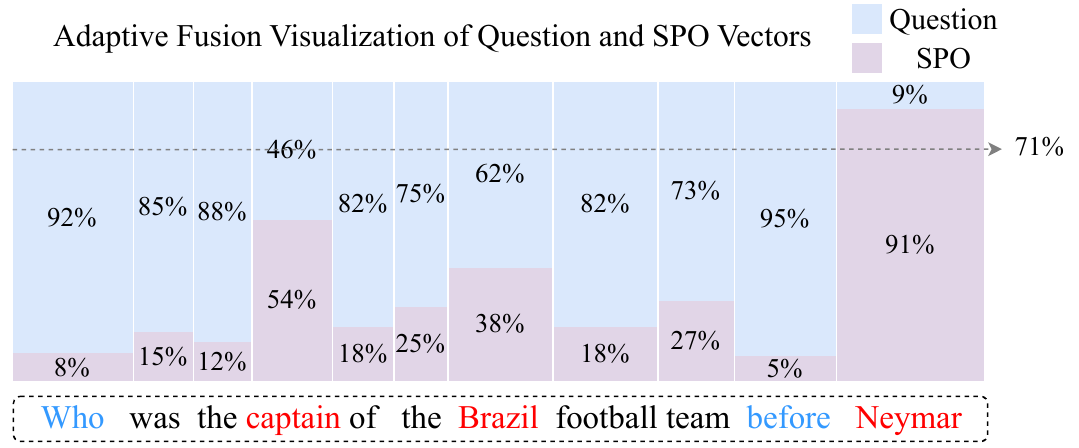}
    \caption{Visualization of the question and SPO vectors using the adaptive fusion technique. 
    %In the horizontal axis, the width of each rectangle represents the word weight in the question. In the vertical axis, the orange color represents the proportion of question information, and the yellow color represents the proportion of SPO information.%Taking the SPO $\langle$Neymar, position held, captained of Brazil national team, 2014$\rangle$ as an example, 
    %On the horizontal axis, the width of each rectangle represents the word weight. On the vertical axis, the orange and yellow represent the proportion of question information and the proportion of SPO information, respectively.
    }
    \label{fusion}
\end{figure}

The results are presented in Table \ref{ablation}. 
We can obtain the following insights:  First, after eliminating the Question Calibration, the model's performance drops drastically, which is in line with our expectations. This result indicates that this module can provide helpful contextual information for accurately understanding the question. 
Second, the performance declines when we perform one-hop message passing instead of multi-hop, empirically demonstrating that multi-hop message passing is more expressive. 
Finally, complex questions require the temporal order information to be captured, thus removing this information inevitably harms the model.
%Second, since the local information can bring additional valuable information from KGs, eliminating it can negatively affect the model. Moreover, the performance declines when we perform one-hop message passing instead of multi-hop, empirically demonstrating that multi-hop message passing is more expressive. Finally, complex questions require the temporal order information to be captured, thus removing this information inevitably harms the model.

\begin{figure}[htbp]
\centering
\includegraphics[width=0.45\textwidth]{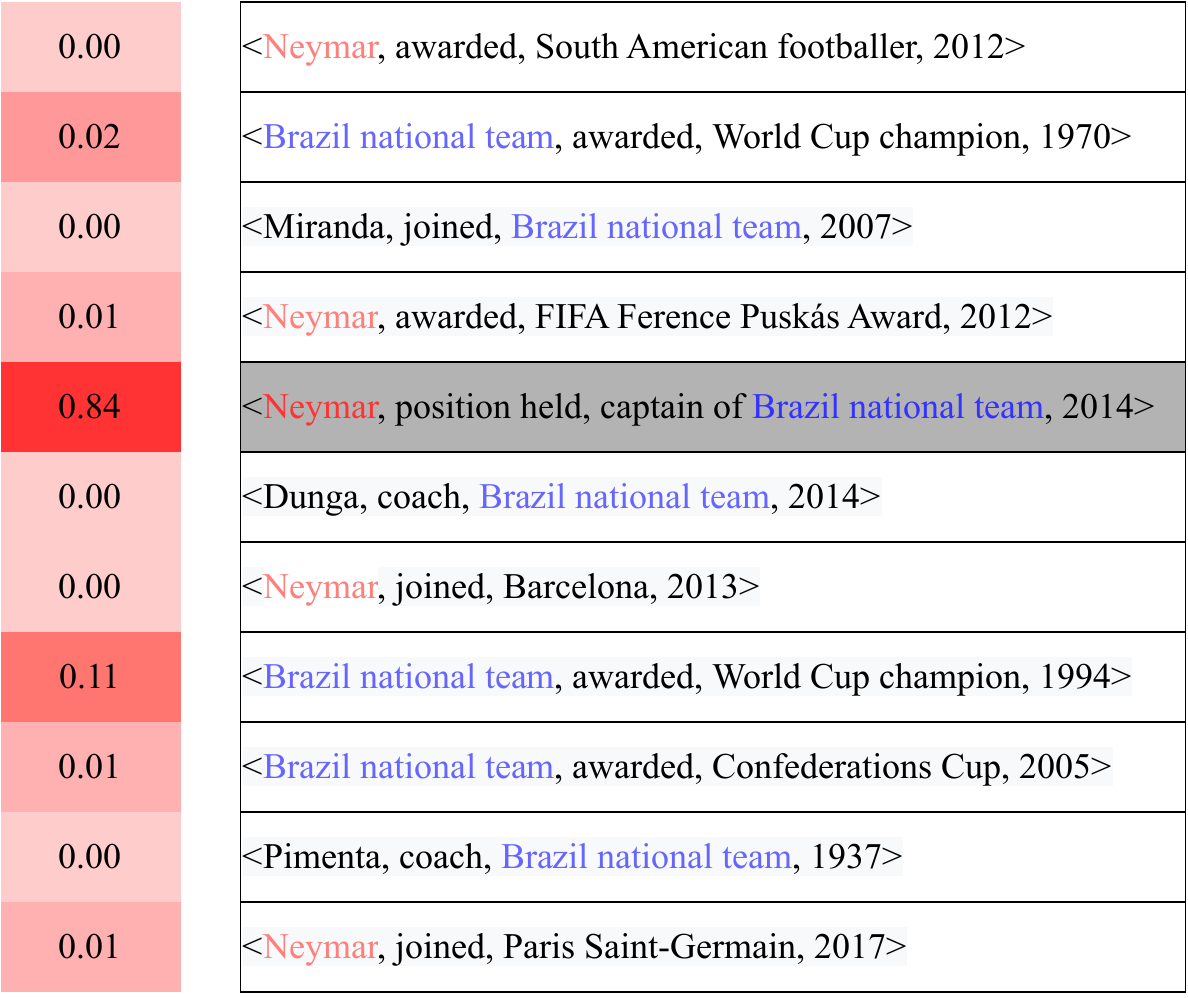}
\caption{Visualization of the SPO gradients. The left if percentage of gradients, and the right is details of the SPOs. 
%Deeper colors imply more significant contributions to the final prediction. The right shows the SPOs' details.
%correspond to a higher proportion of gradients, 
}
\label{spo}
\vspace{-0.2cm}
\end{figure}

\subsection{Interpretability of Multi-Hop Modeling}\label{Interpre of M-h Modeling}
To interpret our model's reasoning process, we investigate the relational path attention weights induced by the attention layer of GNNs described in Eq. \ref{attention}. Specifically, we trace high attention weights from entity nodes to the candidate answer nodes on the retrieved sub-graph $G_q$ by leveraging Best First Search (BFS). Fig. \ref{example} illustrates one example. In this example, we note that the reasoning path contains ``Cristiano Ronaldo'' in the question and ``Alex Ferguson'' and ``Manchester United F.C.'' in KGs.  QC-MHM can make accurate predictions, \textit{i.e.}, ``Alex Ferguson'', given the question. %where ``Alex Ferguson'' is predicted as the final answer. the path of attention weights between entities ``Obama'' and ``President of USA'' drops consistently as increasing the number of layers. On the converse, the attention weights between the correct answer and ``President of USA'' increase all the time. 
Notably, QC-MHM promotes rational reasoning by introducing ``Manchester United F.C.'', which is not mentioned in the question, revealing the importance of background knowledge. 
It provides an interpretable reasonable path ``Cristiano Ronaldo$\rightarrow$Manchester United F.C.$\rightarrow$Alex Ferguson''.

%\subsection{Interpretability of Multi-hop Modeling}
%\label{interpre}

%To our knowledge, we are the first work to consider multi-hop message passing within a single layer in the temporal KGQA task.

\subsection{Interpretability of Question Calibration}
\label{res:IA}

To validate the role of SPO information selection, we use the question ``Who was the captain of the Brazil national team before Neymar?'' for a quantitative study. First, we take the hidden states of each SPO separately and then calculate their gradients. Finally, a normalization function is applied to obtain each SPO proportion. As shown in Fig. \ref{spo}, the more helpful the SPO is in understanding the question, the higher its gradient proportion is., indicating that QC-MHM can capture useful inference information from SPOs. And, To verify the ratio of SPO and question fusion, Fig. \ref{fusion} visualizes each word weight in the question and its information fusion ratio. QC-MHM decreases the information share of the entity word ``Neymar'' and increases the information from SPO. The information distribution is consistent with human cognition, indicating that QC-MHM mitigates the influence of the entities in the question while effectively fusing the SPO information with the question information adaptively.

\section{Conclusion}
\label{con}
In this work, we propose the Question Calibration and Multi-Hop Modeling approach for the temporal KGQA task. Three specific modules are introduced to significantly improve the performance of the model. Specifically, a time-sensitive KG embedding module is used to add temporal ordering information. In addition, the question calibration and multi-hop modeling module adaptively fuse the SPO in the graph and explicitly model the multi-hop question, and finally gets the answer in the answer prediction module. Extensive experiments on two widely used datasets show that QC-MHM achieves consistent improvements over baselines.

\section{Acknowledgments}
This work was supported by the National Key Research and Development Program of China (No. 2021YFB1714300) and National Natural Science Foundation of China (No.62006012, No.62132001)

\bibliography{aaai24}
%\appendix
%\input{appendix}
\end{document}